\documentclass[a4paper,conference]{IEEEtran}
\ifCLASSINFOpdf
\else
\fi

\usepackage{graphics} 
\usepackage{epsfig} 
\usepackage{mathptmx} 
\usepackage{times} 
\usepackage{amsmath} 
\usepackage{amssymb}  

\begin{document}
%
\title{Attributes Aware Face Generation with Generative Adversarial Networks}



%
\author{\IEEEauthorblockN{Zheng Yuan\IEEEauthorrefmark{1}\IEEEauthorrefmark{2},
Jie Zhang\IEEEauthorrefmark{1},
Shiguang Shan\IEEEauthorrefmark{1}\IEEEauthorrefmark{2},
Xilin Chen\IEEEauthorrefmark{1}\IEEEauthorrefmark{2}}
\IEEEauthorblockA{\IEEEauthorrefmark{1}Institute of Computing Technology, Chinese Academy of Sciences, China}
\IEEEauthorblockA{\IEEEauthorrefmark{2}University of Chinese Academy of Sciences, Beijing, China}}


\maketitle

\begin{abstract}
  Recent studies have shown remarkable success in face image generations. However, most of the existing methods only generate face images from random noise, and cannot generate face images according to the specific attributes. In this paper, we focus on the problem of face synthesis from attributes, which aims at generating faces with specific characteristics corresponding to the given attributes. To this end, we propose a novel attributes aware face image generator method with generative adversarial networks called AFGAN. Specifically, we firstly propose a two-path embedding layer and self-attention mechanism to convert binary attribute vector to rich attribute features. Then three stacked generators generate $64 \times 64$, $128 \times 128$ and $256 \times 256$ resolution face images respectively by taking the attribute features as input. In addition, an image-attribute matching loss is proposed to enhance the correlation between the generated images and input attributes. Extensive experiments on CelebA demonstrate the superiority of our AFGAN in terms of both qualitative and quantitative evaluations.
\end{abstract}


%
\IEEEpeerreviewmaketitle

\section{Introduction}
\begin{figure*}[htbp]
  \centering
  \includegraphics[width=1\textwidth]{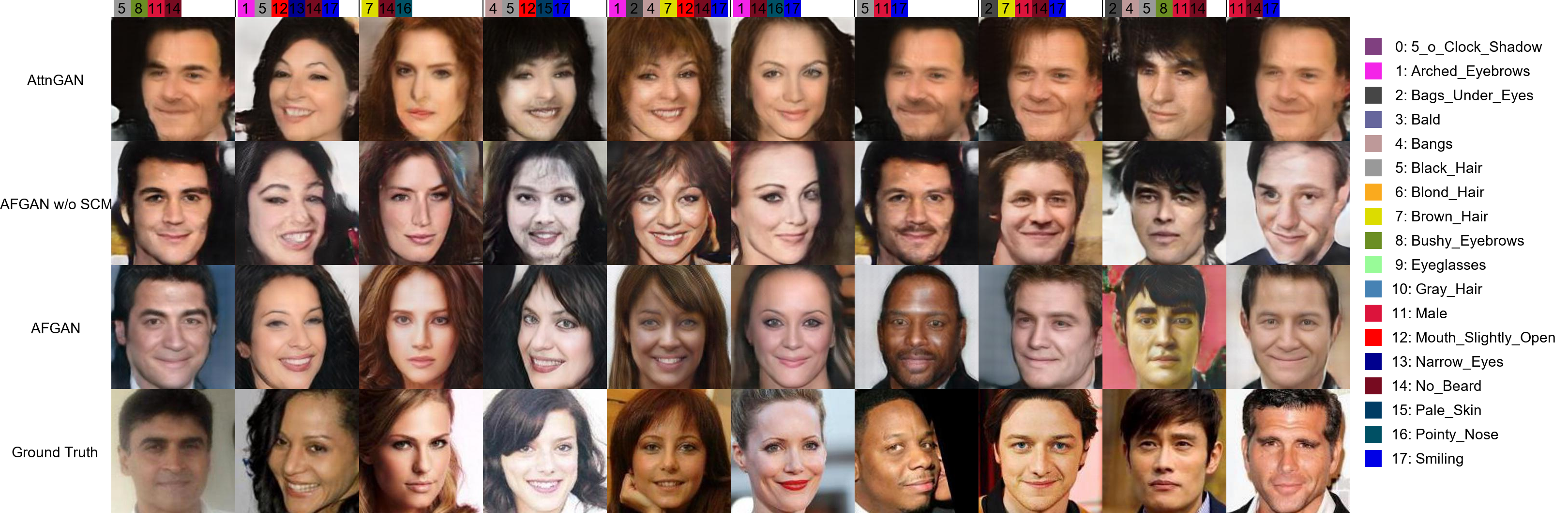}
  \vspace{-2ex}
  \caption{The generated face images in ablation study of AFGAN model. The first row represents the results of AttnGAN~\cite{xu2018attngan}, 
  second row represents the experiment results without the constraints of SCM module in the objective function, and third row represents the results of the complete
  AFGAN model proposed by us. The input attribute label is shown in the top color block and number, 
  and the corresponding attribute content is listed on the right side. The ground truth images corresponding to the attribute labels are 
  shown in the last row.}
  \label{fig:merge_four}
  \vspace{-3ex}
\end{figure*}

Recently, more and more attention has been paid to image generation. Great progresses 
have been achieved by generative adversarial networks and its variants~
\cite{goodfellow2014generative, mirza2014conditional}. Different from the general 
image generation, the face synthesis pays more attention on the details of the generated face. Our work aims at generating face images with specific input attributes, i.e., attributes aware face generation, which has not received a wide range of 
attention in the past. Face generation with specific attributes has wide application 
prospects. For example, it can be utilized to extend face datasets for improving the 
face recognition models or provide a face synthesis of the suspect for criminal 
investigations.

Nowadays, there are several breakthroughs both in diversity and clarity for face 
generation. The StyleGAN~\cite{karras2018style} and its upgraded version StyleGANv2~\cite{karras2019analyzing}, as one of the state-of-the-art 
methods in face generation, can generate a high resolution face image of $1024\times1024$ 
pixels. The generated high resolution image contains detailed face information, e.g., 
the hairstyle, the thickness of eyebrows, the beard type, etc. However, the 
attributes of faces are randomly generated and a face with some specific attributes 
can not be achieved.

Text-to-image generation is another work related to attributes aware face generation. 
AttnGAN~\cite{xu2018attngan}, as one of state-of-the-art methods, generates images through 
a stacked generator and multimodal similarity module. The image generated by AttnGAN is closely related to the input text information and 
can well characterize the scene described by the text content. However the input 
attribute labels are different from the text sentence, it is non-trivial to directly 
apply AttnGAN for attributes aware face generation. 
DM-GAN~\cite{zhu2019dm}, Obj-GAN~\cite{li2019object} and OP-GAN~\cite{hinz2019semantic} are recently proposed to further solve the task of text-to-image generation. Due to the dynamic memory mechanism and image layout information used in these works, the generated image is more consistent with the input text.

In the task of attribute-driven face image generation, which are closely related to our work, \cite{lu2018attribute, yan2016attribute2image, wang2018attribute} are proposed to solve this problem, but the resolution and clarity of the generated images still need to be improved.

In this paper, we propose a novel framework named Attributes Aware Face Generation 
with Generative Adversarial Networks (AFGAN) for generating face images from the 
input attributes. The model is mainly composed of three modules: attribute embedding 
module, stacked image generation module and similarity constrain module. In attribute 
embedding module, we propose a two-path embedding layer to convert the input attribute 
vector into rich face attribute feature, which is different from the 
naive embedding table in the word embedding layer from AttnGAN~\cite{xu2018attngan}. 
Since in our task, whether the value of the attribute is 0 or 1, the input of face 
attribute always has certain meaning. Such as "young" attribute, when the value of 
this attribute is 1, it means young people and when the value is 0, it indicates 
old people. So we design a two-path embedding layer to make the input attributes 
well reflect their meanings, which benefits to the subsequent image generation. 
At the same time, we also use the self-attention module~\cite{zhang2018self} right 
after the embedding layer to introduce interconnections between different attributes. 
In stacked image generation module, we use a stacked three-level image generator 
to generate images of $64 \times 64$, $128 \times 128$ and $256 \times 256$ pixels respectively. On the one hand, the 
coarse-to-fine framework of image generation can reduce the learning burden of 
image generator. On the other hand, it can gradually improve the quality of the 
generated image, so as to we achieve a clear and realistic image which 
is consistent with the input face attributes. In the similarity constrain module, 
the high-level face feature vector is firstly extracted from the $256\times 256$ 
face image through a pre-trained Inception\_v3 model of face attributes 
prediction. Then the high-level face feature vector and the face attribute 
feature from the two-path embedding layer are transformed into the same feature space. By minimizing the distance 
between the two feature vectors, the generated face image can well reflect the 
input face attributes.

The contributions of our work are summarized as follow:
\begin{itemize}
\item We propose a novel framework named AFGAN for generating face images from face attributes.
\item A two-path embedding layer of face attributes is proposed to well characterize the attributes for face generation. The attributes information can be accurately transmitted to the subsequent image generation whether the value of each face attribute is 0 or 1.
\item Both qualitative and quantitative experiment results show that the face images generated by our AFGAN can not only conform to the input face attributes, but also have facial details with good image quality and clarity.
\end{itemize}

\section{Related Work}
Our work is aimed at generating face images according to the input attributes. 
We expect that the generated images can fully reflect the characteristic of the 
corresponding attributes. Limited to the scope of our study, the most related 
works are high-resolution face image synthesis, text-to-image generations, face hallucination and attribute-driven face image synthesis.

\subsection{High-resolution face image synthesis}
Face image synthesis, as one typical topic of image synthesis, has received 
wide attentions since the proposal of generative adversarial networks~\cite{goodfellow2014generative}. Different from the conventional image generation 
task, which pays more attention to the general structure and the shape of object 
in the generated image, the task of face image synthesis focuses on the details 
of the face, such as the texture of the hair, facial wrinkles and skin gloss, 
etc. Recently there have been several state-of-the-art methods~
\cite{karras2017progressive, karras2018style}, which focus on generating 
high-resolution realistic faces. PGGAN~\cite{karras2017progressive} proposes a 
novel framework which gradually upsamples the generated images from 
$4 \times  4$ to $1024 \times  1024$. The network structure of model is gradually 
deepened, starting at a resolution of $4 \times  4$, and gradually double the 
resolution of the image by adding an upsampling module. Finally PGGAN can generate high-quality faces 
with $1024 \times  1024$ pixels. StyleGAN~\cite{karras2018style}, as one 
state-of-the-art work of face generation, adds a hidden vector mapping module 
based on PGGAN, which embeds the original noise to hidden vectors via eight 
fully connected layers. Then the generated hidden vectors utilized as 
AdaIN~\cite{huang2017arbitrary} factor are fed to each layer for different 
resolution image generation. At the same time, StyleGAN also introduces noises 
into each layer to increase the diversity of the generated face 
images from coarse to fine. Although the faces generated from these works are 
realistic and some of them can mix the spurious with the genuine, we can not 
achieve faces according to specific attributes.

\subsection{Text-to-image generation}
Text-to-image generation takes the text description as input and generates the 
corresponding image which is consistent with the semantic of text. AttnGAN~
\cite{xu2018attngan} uses a stacked three-stage structure to generate images 
from $64\times64$ to $256\times256$ resolution gradually. And a DAMSM module is proposed to 
constrain the distance between the text and generated image by the method of 
encoding image and text to the same semantic space with attention mechanism~
\cite{zhang2018self}. MirrorGAN~\cite{qiao2019mirrorgan} draws on the ideas 
of AttnGAN~\cite{xu2018attngan} and CycleGAN~\cite{zhu2017unpaired} at the 
same time. The generated image is encoded to text again and constrains the 
generated text as similar as the input text.

The methods mentioned above can well generate images containing some objects 
and attributes described by text, but the generated images are mainly focused 
on general objects like flowers and birds rather than faces. The face 
generation should pay more attention to the details of the texture of the hair, 
facial wrinkles and skin gloss, etc., which may be more difficult to generate 
than general objects. Moreover, generation by attribute labels is also different 
from that by text descriptions. Our proposed AFGAN takes the attribute vectors 
as input instead of the text descriptions, which focuses on the interrelations 
between different attributes. Our method can well generate faces with specific 
attributes which ensures both the clarity and authenticity.

\subsection{Face hallucination}
Some works of face hallucination are also related to attribute-guided face image synthesis.
Yu et al. ~\cite{yu2019semantic} uses an attribute-embeded upsampling network to generate high resolution images from tiny unaligned face images, which reducing the uncertainty of one-to-many mappings remarkably.
Li et. al. ~\cite{li2019deep} constructs a face transfer network, which upsamples low-resolution face images to high-resolution images by fusing facial attributes.
The difference between face hallucination and our attribute-guided face images synthesis is that our approach has no input images as reference.

\subsection{Attribute-driven face image synthesis}
There are also some works focusing on attribute-driven face image synthesis, and we give discussions on differences between them and ours.

Attribute2sketch2face~\cite{Di2017Face} firstly synthesizes the facial sketch 
corresponding to the input attributes and then reconstructs the face image 
based on the synthesized sketch. However, the resolution of the generated face 
image is limited to $64  \times  64$. And this work needs the facial sketch as 
the input. Differently our method designs a stacked three-stage generator 
to achieve a high resolution facial image by only taking the attribute label 
as input.

For Lu et al.~\cite{lu2018attribute}, two variants of CycleGAN are proposed to generate attribute-guided images and identity-guided images respectively. This method takes low resolution faces as input, and combines attributes to generate high-resolution face images. Differently, our AFGAN does not utilize any faces as input, but purely generates face images from attributes. 

Yan et al. ~\cite{yan2016attribute2image} considers that an image is a combination of foreground and background. A VAE framework is employed to generate face image through disentangled latent variables. However, this work only generates 64 $\times$ 64 images, and our work focuses on how to generate higher resolution images, i.e., 256 $\times$ 256. Besides, the image generated by Yan et al. ~\cite{yan2016attribute2image} looks fuzzy and lacks of diversity while our AFGAN can generates higher quality images.

For Wang et al.~\cite{wang2018attribute}, a DCGAN-based model is proposed to generate face images by adding attribute vector to the input. At the same time, generation of continuous sequence of face images is also considered in this work. However, the generated images shown in ~\cite{wang2018attribute} are fuzzy, and our method outperforms~\cite{wang2018attribute} in terms of both IS and FID metrics. Moreover, Wang et al.~\cite{wang2018attribute} only focuses on five basic attributes, e.g. glasses, gender, hair color, smile and age, while our work focuses on 18 attributes, including more attributes with some particular ones, like pointy nose, bushy eyebrows and so on.

However, the resolution of images generated by these works ranges from 64 to 128, and the image size of our proposed method is 256. So the quality and clarity of the images generated by our method are much better than these works.

\section{Approach}
In this section, we will start with the overview of our method, 
and then introduce the three modules respectively. Finally we give the 
objective function of the whole model.

\subsection{Overview}
\begin{figure*}[thpb]
  \centering
  \includegraphics[width=0.9\textwidth]{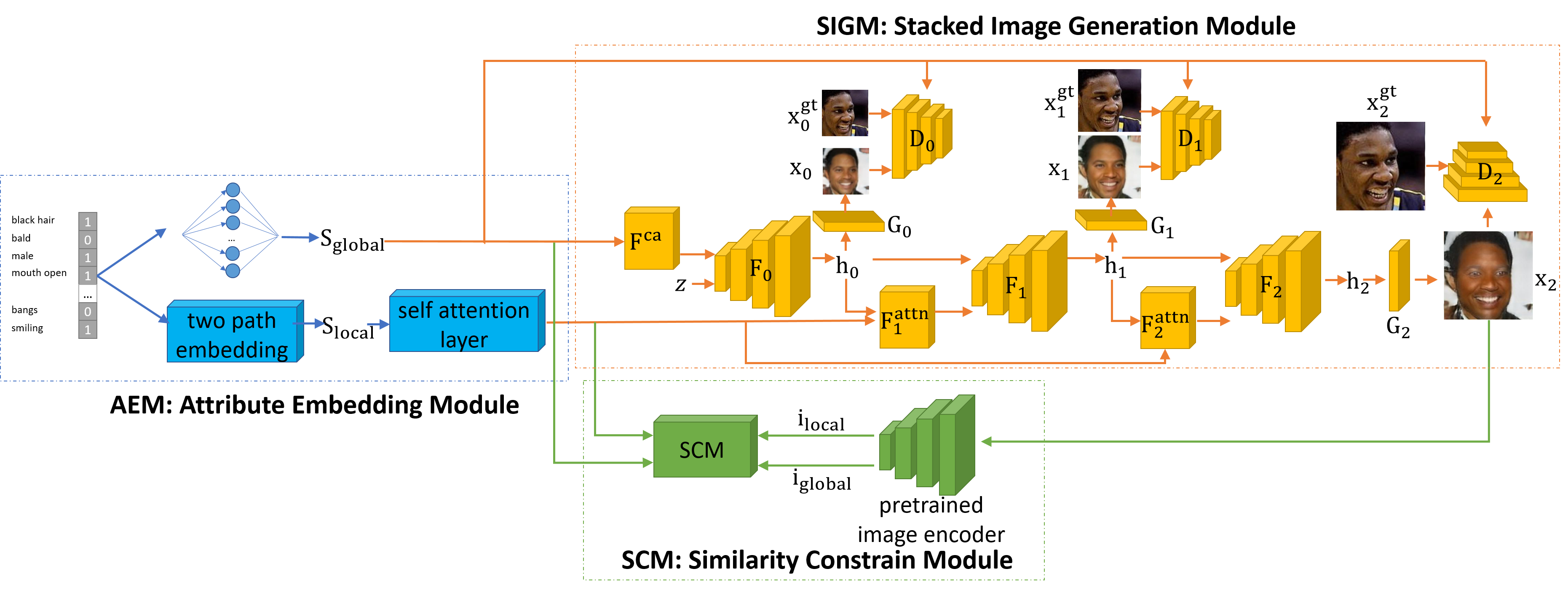}
  \vspace{-3ex}
  \caption{The structure of AFGAN. The model consists of three modules: attribute embedding module (AEM), stacked image generation module (SIGM) and similarity constrain module (SCM). 
  In AEM, $S_{global}$ and $S_{local}$ represent the global and local features of input attributes. 
  In SIGM, $z$ is the noise vector. $F^{ca}$ denotes conditioning augmentation module~\cite{zhang2017stackgan}. $F_{i}^{attn}$, $F_i$, $G_i$ and $D_i$ are attention module, upsampling block, generation module and discriminator in each stage respectively. $h_i$ is the hidden state vector transmitted in different stages. $x_i$ and $x_i^{gt}$ are the generated and corresponding ground truth images in each stage.
  In SCM, $i_{local}$ and $i_{global}$ represent the local and global image features extracted by a pretrained image encoder.}
  \label{fig:AFGAN}
  \vspace{-4ex}
\end{figure*}

As shown in Figure \ref{fig:AFGAN}, AFGAN consists of three modules: attribute 
embedding module (AEM), stacked image generation module (SIGM) and similarity 
constrain module (SCM). AFGAN takes the attribute vector as input. In the AEM, 
the attribute vector is converted to global attribute features and local 
attribute features through the two-path embedding layer and self-attention 
layer. In the SIGM, the global attribute feature firstly uses conditioning 
augmentation method~\cite{zhang2017stackgan} to increase the diversity of the 
input attribute. Then it is utilized to generate face images through stacked image 
generator, from low resolution to high resolution gradually together with the 
local attribute feature. In the SCM, we first encode the generated images 
through a pretrained inception\_v3~\cite{szegedy2016rethinking} model to 
extract high-level features. Then we convert the encoded image feature and 
local attribute feature into the same semantic space. Through constraining the 
distance of two features by extra objective function term, we aim at forcing 
the generated face images to keep pace with the input face attribute.
\subsection{AEM: Attribute Embedding Module}
We denote the input attribute vector as $S_{attr} \in \{0, 1\}^N$, $N$ is the 
number of input attributes. We first convert it into $S_{global} \in \mathbb{R}^C$, 
which represents the global semantics of attribute vectors and 
$S_{local} \in \mathbb{R}^{N \times C}$, which represents the semantics of each 
attribute. $C$ is the dimension of feature space. Through different levels of 
semantic vectors, we hope that both the high-level semantic information and 
low-level detailed information existing in the input attribute label can be 
transmitted during the process of image generation. Specifically, $S_{global}$ 
is transformed through a fully connected layer:
\begin{equation}
  \setlength{\abovedisplayskip}{3pt}
  \setlength{\belowdisplayskip}{3pt}
  S_{global} = W_{global} * S_{attr},
\end{equation}
\begin{figure}[thbp]
  \centering
  \includegraphics[scale=0.5]{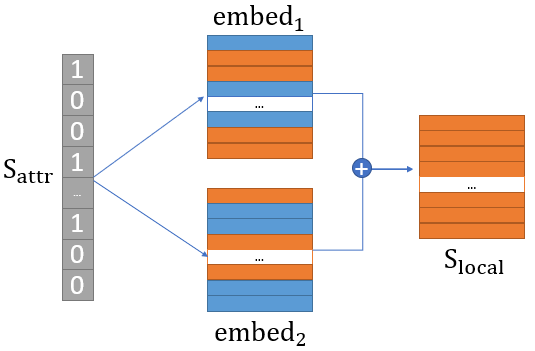}
  \vspace{-2ex}
  \caption{The structure of two path embedding layer. $S_{attr}$ is the one-hot attribute vector. $embed_1$ and $embed_2$ are two embedding tables, representing the semantics of each input attribute whose value is 0 and 1 respectively. $S_{local}$ represents the semantics of each attribute. The orange vectors in the figure mean the activated vectors.}
  \label{fig:two_path_embed}
  \vspace{-2ex}
\end{figure}
where $W_{global} \in \mathbb{R}^{C \times N}$. $S_{local}$ is transformed by 
our proposed two-path embedding layer, as shown in Figure 
\ref{fig:two_path_embed}, and the process can be formulated as the following:
\setlength{\abovedisplayskip}{3pt}
\setlength{\belowdisplayskip}{3pt}
\begin{equation}
S_{local} = embed_1 * S_{attr} + embed_2 * (1-S_{attr}),
\end{equation}
where $embed_1, embed_2 \in \mathbb{R}^{C \times N}$ are two embedding tables, 
representing the semantics of each input attribute whose value is 0 and 1 
respectively. The reason behind this design is that in our work, whether the 
value of the attribute is 0 or 1, the input of face attribute always has 
certain meaning. Such as "young" attribute, when the value of this attribute 
is 1, it means young people while 0 indicates old people. 
The carefully designed two-path embedding layer can generate feature vector 
which well reflects their meanings of the input attribute.

Next, we use self-attention layer~\cite{zhang2018self} to focus on modeling the 
relationships between different attributes:
\vspace{-1ex}
\begin{gather}
  \setlength{\abovedisplayskip}{3pt}
  \setlength{\belowdisplayskip}{3pt}
f(x) =W_{f} * S_{local}, \quad g(x) =W_{g} * S_{local}, \\ 
s_{i j} =f\left(x_{i}\right)^{T} g\left(x_{j}\right), \quad \beta_{i, j} =\frac{\exp \left(s_{i j}\right)}{\sum_{i=1}^{N} \exp \left(s_{i j}\right)}, \\
h(x) =W_{h} * S_{local}, \quad S_{local'_j} =\sum_{i=1}^{N} \beta_{i, j} h\left(x_{i}\right), \\ 
S_{local'} =\left(S_{local'_1}, S_{local'_2}, \ldots, S_{local'_N}\right) \in \mathbb{R}^{C \times N}, 
\end{gather}
where $W_f$ and $W_g$ are query and key matrix respectively, used to calculate the correlation 
between the different attributes, and $W_h$ is value matrix. Through self-attention layer, we 
can achieve enhanced attribute features $S_{local'}$ which taking into account the relationship between different
 attributes.

\subsection{SIGM: Stacked Image Generation Module}
Stacked image generation module is composed of three stacked image generation 
modules. Firstly, an image of $64 \times 64$ pixels is generated, and then the scale of 
the generated image is gradually increased to $128 \times 128$. Finally the $128 \times 128$ pixels 
image is transitioned to a high resolution version of $256\times256$ pixels. The three 
generation modules are respectively recorded as $G_0$, $G_1$ and $G_2$, and 
their corresponding hidden state vectors transmitted between different 
generations are recorded as $h_0$, $h_1$, $h_2$. The corresponding generated 
images are recorded as $x_0$, $x_1$, $x_2$. The whole process can be formulated 
as follow:
\begin{align}
  \setlength{\abovedisplayskip}{3pt}
  \setlength{\belowdisplayskip}{3pt}
h_{0} &=F_{0}\left(z, F^{c a}(S_{global})\right), \\ 
h_{i} &=F_{i}\left(h_{i-1}, F_{i}^{attn}\left(S_{local'}, h_{i-1}\right)\right) \text { for } i=1,2, \\ 
x_{i} &=G_{i}\left(h_{i}\right),
\end{align}
where $z$ is the noise vector sampled from the standard normal distribution, $F^{ca}$ denotes conditioning augmentation module~\cite{zhang2017stackgan} used for converting global attribute vector $S_{global}$ to diverse conditional vector, $F_{i}^{attn}$ is the attention module used in stage $i$.

Attention module $F^{attn}$ is utilized to calculate the attribute-content matrix in each stage of image generation, which represents the relationship between each attribute of the input and each region in the generated image. $F^{attn}$ has two inputs: local attribute feature $S_{local'}$ and the hidden state vector from the previous layer. The attribute-content matrix is calculated as follow:
\begin{gather}
  \setlength{\abovedisplayskip}{3pt}
  \setlength{\belowdisplayskip}{3pt}
s_{j, i}^{\prime} =h_{j}^{T}S_{local_i}, \quad \beta_{j, i} =\frac{\exp \left(s_{j, i}^{\prime}\right)}{\sum_{k=1}^{N} \exp \left(s_{j, k}^{\prime}\right)}, \\ 
F^{attn}(S_{local}, h)=\left(c_{1}, c_{1}, \ldots, c_{N}\right) \in \mathbb{R}^{{D} \times N}, \quad
c_{j} =\sum_{i=1}^{N} \beta_{j, i} S_{local_i}.
\end{gather}

At the same time, each generation module corresponds to a discriminator module $D_i$, which is used to judge the quality of the generated image. Each discriminator consists of an unconditional discriminator and a conditional discriminator. The input of the former is only an image, which focusing on the authenticity of the generated image, and the inputs of the latter are both the image and the corresponding attribute vector. It is used to judge the matching degree between the generated image and the input attribute.

\subsection{SCM: Similarity Constrain Module}
Similarity constrain module is used to constrain the generated images to be more consistent with 
the input attributes. We first encode the generated images with a pretrained inception\_v3~
\cite{szegedy2016rethinking} model, which is trained as an attribute predictor, to achieve high 
level feature of the image. Specifically, we use the feature from "mixed\_6e" layer as local 
image feature $i_{local} \in \mathbb{R}^{768\times289}$, where 768 is the feature dimension and 
289$(17*17)$ denotes the number of regions in the image. And we use the feature from the last 
global pooling layer as the global image feature $i_{global} \in \mathbb{R}^{2048}$.

We use the local attribute feature $S_{local'}$ and local image feature $i_{local}$ to calculate 
the matching degree between the generated image and input attribute in terms of local evaluation:
\begin{equation}
s=S_{local'}^{T} i_{local} \in \mathbb{R}^{N \times 289}, \quad \overline{s}_{i, j}=\frac{\exp \left(s_{i, j}\right)}{\sum_{k=1}^{N} \exp \left(s_{k, j}\right)},
\end{equation}
\begin{equation}
\alpha_{i,j}=\frac{\exp \left(\gamma_{1} \overline{s}_{i, j}\right)}{\sum_{k=1}^{289} \exp \left(\gamma_{1} \overline{s}_{i, k}\right)}, \quad c_{i}=\sum_{j=1}^{289} \alpha_{i,j} i_{local_j},
\end{equation}
\begin{equation}
R\left(c_{i}, S_{local'_i}\right)=\frac{c_{i}^{T} S_{local'_i}}{\left\|c_{i}\right\|\left\|S_{local'_i}\right\|},
\end{equation}
\begin{equation}
R^{local}(Q, D)=\log \left(\sum_{i=1}^{N} \exp \left(\gamma_{2} R\left(c_{i}, S_{local'_i}\right)\right)\right)^{\frac{1}{\gamma_{2}}},
\end{equation}
where $\gamma_1$ and $\gamma_2$ are hyperparameters.
The global attribute feature $S_{global}$ and global image feature $i_{global}$ are employed to 
calculate another matching degree between the generated image and input attribute in terms of 
global evaluation:
\begin{equation}
R^{global}(Q, D)=\frac{i_{global}^{T} S_{global}}{\|i_{global}\|\|S_{global}\|}.
\end{equation}

This two matching degrees can be used to evaluate the quality of the generated image and reflect 
whether it matches the input attributes.

\subsection{Objective function}
The objective function of the whole network is defined as follows:
\vspace{-3ex}
\begin{equation}
  \mathcal{L}=\mathcal{L}_{G}+\lambda \mathcal{L}_{SCM}, \quad \mathcal{L}_{G}=\sum_{i=0}^{2} \mathcal{L}_{G_{i}},
\end{equation}
where $\mathcal{L}_{G}$ denotes the objective functions from three stacked image generation modules and $\mathcal{L}_{SCM}$ denotes the objective function in SCM module.

The adversarial loss function for $G_i$ is defined as follows:
\vspace{-1ex}
\begin{equation}
\mathcal{L}_{G_{i}}=-\frac{1}{2} \mathbb{E}_{x_{i} \sim p_{G_{i}}}\left[\log \left(D_{i}\left(x_{i}\right)\right)\right]-\frac{1}{2} \mathbb{E}_{x_{i} \sim p_{G_{i}}}\left[\log \left(D_{i}\left(x_{i}, S_{global}\right)\right)\right].
\end{equation}

The adversarial loss function for $D_i$ is defined as follows:
\begin{equation}
\begin{aligned}
		\mathcal{L}_{D_{i}}=&-\frac{1}{2} \mathbb{E}_{x_{i}^{gt} \sim p_{data_{i}}}\left[\log D_{i}\left(x_{i}^{gt}\right)\right]-\frac{1}{2} \mathbb{E}_{x_{i} \sim p_{G_{i}}}\left[\log \left(1-D_{i}\left(x_{i}\right)\right)\right] \\
		&-\frac{1}{2} \mathbb{E}_{x_{i}^{gt} \sim p_{data_i}}\left[\log D_{i}\left(x_{i}^{gt}, S_{global}\right)\right]\\
		&-\frac{1}{2} \mathbb{E}_{x_{i} \sim p_{G_{i}}}\left[\log \left(1-D_{i}\left(x_{i}, S_{global}\right)\right)\right],
\end{aligned}
\end{equation}
where $x_i^{gt}$ is the ground truth image in the corresponding generator.

For the objective function in SCM module, we first use the local matching degree $R^{local}(Q, D)$ to calculate the matching degree between the generated image and input attribute for each input attribute in a minibatch of training data:
\begin{equation}
P\left(D_{i} | Q_{i}\right)=\frac{\exp \left(\gamma_{3} R^{local}\left(Q_{i}, D_{i}\right)\right)}{\sum_{j=1}^{M} \exp \left(\gamma_{3} R^{local}\left(Q_{i}, D_{j}\right)\right)},
\end{equation}
where $\gamma_3$ is a hyperparameter and $M$ is the number of samples in a minibatch of training data.
And the corresponding objective function can be defined as follows:
\begin{equation} 
  \mathcal{L}_{1}^{local}=-\sum_{i=1}^{M} \log P\left(D_{i} | Q_{i}\right). 
\end{equation}

Similarly, we can defined the matching degree between the generated image and input attribute for each generated image in a minibatch of training data, and calculate the corresponding objective function:
\begin{align}
P\left(Q_{i} | D_{i}\right)&=\frac{\exp \left(\gamma_{3} R^{local}\left(Q_{i}, D_{i}\right)\right)}{\sum_{j=1}^{M} \exp \left(\gamma_{3} R^{local}\left(Q_{j}, D_{i}\right)\right)}, \\
\mathcal{L}_{2}^{local}&=-\sum_{i=1}^{M} \log P\left(Q_{i} | D_{i}\right).
\end{align}

Similarly by using the global matching loss $R^{global}(Q, D)$, we can correspondingly achieve $\mathcal{L}_{1}^{global}$ and $\mathcal{L}_{2}^{global}$. Finally, the objective function of SCM is defined as follows:
\begin{equation}
  \label{for:SCM_loss}
  \mathcal{L}_{SCM}=\mathcal{L}_{1}^{local}+\mathcal{L}_{2}^{local}+\mathcal{L}_{1}^{global}+\mathcal{L}_{2}^{global}.
\end{equation}

\section{Experiment}
In this section, we will demonstrate the superiority of our AFGAN through both qualitative and quantitative evaluations. We first introduce the experiments setting and the implementation details of our method. 
																									  
Then we present the ablation study of our method to investigate the effectiveness of each module proposed 
in AFGAN. Furthermore, we compare AFGAN with the state-of-the-art method AttnGAN~\cite{xu2018attngan} in terms of qualitative evaluations. Finally we 
employ common metrics (e.g. BRISQUE, IS, FID and MS-SSIM) and an extra trained facial attribute predictor to do quantitative evaluations.

\subsection{ Experimental Settings}
					   
We use CelebA~\cite{liu2015faceattributes} dataset to evaluate our proposed AFGAN. CelebA dataset consists of 202599 face images, and each has 40 kinds of facial attributes. After performing face detection on all faces, 180694 faces are treated as the training set and 19761 faces are used for testing. Since some attributes can not be recognized from a tight face region (such as necklaces) or some attributes are difficult to measure (such as attractive), we select 18 attributes from them for our experiments. The selected 18 attributes are listed in Table \ref{tab:celeba_attr}.

We compare our method with the state-of-the-art method AttnGAN\cite{xu2018attngan} in terms of both qualitative and quantitative evaluations. Since the AttnGAN is a text-to-image model, and the input of our work is attribute vector instead of text, we change the onehot text vector into onehot attribute vector as the input of model.

\begin{table}
\caption{18 attributes selected from CelebA dataset.}
\vspace{-3ex}
\label{tab:celeba_attr}
    \begin{center}
    \begin{tabular}{|c|c|c|c|}
        \hline
        No. & Attribute & No. & Attribute \\
        \hline
	    0  & 5\_o\_Clock\_Shadow   & 9  & Eyeglasses   \\   
	    1  & Arched\_Eyebrows      & 10 & Gray\_Hair       \\
        2  & Bags\_Under\_Eyes     & 11 & Male         \\
        3  & Bald                  & 12 & Mouth\_Slightly\_Open       \\
        4  & Bangs                 & 13 & Narrow\_Eyes         \\           
        5  & Black\_Hair           & 14 & No\_Beard         \\
        6  & Blond\_Hair           & 15 & Pale\_Skin         \\  
        7  & Brown\_Hair           & 16 & Pointy\_Nose         \\
        8  & Bushy\_Eyebrows       & 17 & Smiling        \\
        \hline
    \end{tabular}
  \end{center}
  \vspace{-5ex}
\end{table}
\subsection{ Implementation details}
Our proposed AFGAN consists of three modules: AEM, SIGM and SCM. 

We first pretrain the image encoder (as shown in Figure \ref{fig:AFGAN}) in SCM. The SCM module is used to calculate the similarity between the input attribute features and the generated image features. Specifically, the input attributes features consist of the global and local features obtained from the AEM module, while the generated image features are obtained by AlexNet, which is a pretrained attribute prediction model. The local features of the generated image are the features before the third pooling layer in AlexNet, and the global features of the generated image come from the last full connection layer. After getting the global and local features of attributes and images respectively, the loss function is calculated according to formula~(\ref{for:SCM_loss}).

The AEM module is composed of global and local parts. The global part is to extract the feature of the input attribute vector through a full connection layer. The local part is mapped by two embedding tables first, as shown in Figure \ref{fig:two_path_embed}. Then a self-attention layer is carried out by using the formula of $softmax (A A^T) A$, where $A$ is the feature generated from the previous two-path embedding layer, so that the relationship between different attributes can be integrated.

Traditional convolution network is employed as discriminators in the first two stages in SIGM to discriminate the authenticity of generated facial 
images. And we use PatchGAN~\cite{isola2017image} as discriminator in the third stage since it can 
pay more attention to details of the generated images. We adopt WGAN\_GP~\cite{gulrajani2017improved} 
loss as the GAN loss. 

We use Adam optimizer~\cite{kingma2014adam} with $\beta_1 = 0.5$ and $\beta_2=0.999$ for training the network. We set the attribute feature dimension
$C$ as 256 and the dimension of noise vector $z$ as 100. We adjust the update frequency of generator and discriminator during the 
training of GAN, i.e., updating once discriminator every four times of generator. 
We train AFGAN model with 30 epochs on 3 GTX 1080Ti GPUs for about 5 days.

\subsection{ Ablation Study }
To demonstrate the superiority of our AFGAN, we do ablation study to investigate the 
effectiveness of each module proposed in AFGAN. 
\subsubsection{ SCM Module}
\begin{figure*}[htbp]
      \centering
      \includegraphics[width=0.8\textwidth]{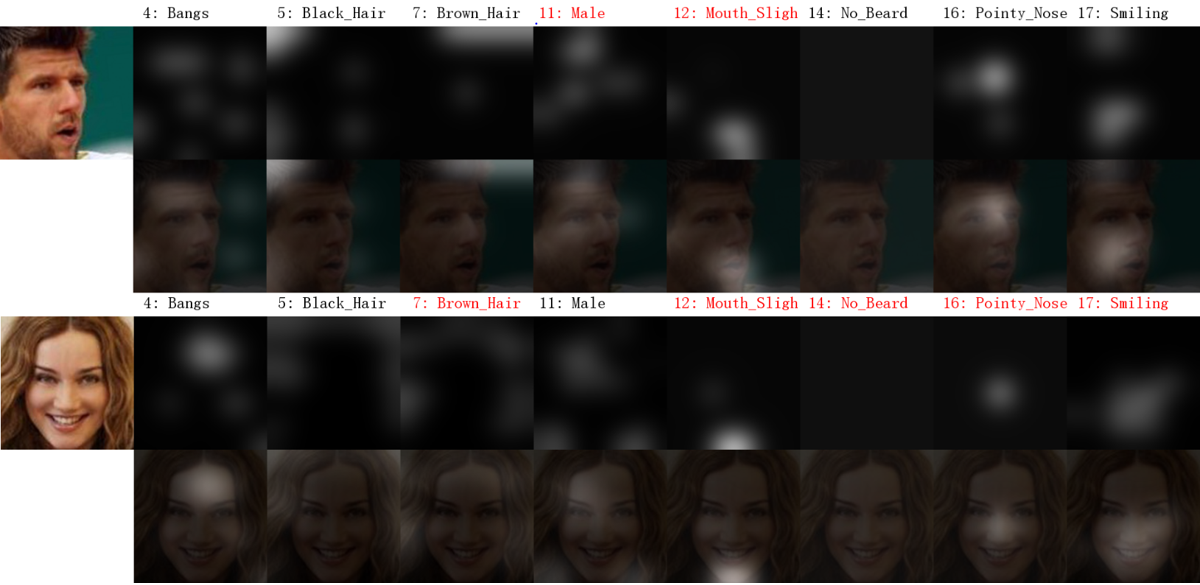}
      \vspace{-2ex}
      \caption{The attention maps in SCM module. The red label means the value of the attribute is 1 and the black 
      means the value of the attribute is 0. The white region of the attention map represents 
      the response area corresponding to the attribute.}
      \label{fig:AFGAN_SCM}
      \vspace{-3ex}
\end{figure*}
																															  
The SCM module uses the pretrained encoder to constrain the distance between the features of 
attribute vector and the generated image. After the training of the SCM module, $\alpha$ in 
SCM represents the correlation between the image and each of input attributes. For each facial 
image, the attention map for the response of each attribute is shown in Figure 
\ref{fig:AFGAN_SCM}, where the red label means the value of the attribute is 1 and the black 
means the value of the attribute is 0. The white region of the attention map represents 
the response area corresponding to the attribute. We can see that the response areas of most 
attributes are reasonable, such as hair (black hair, brown hair), nose, etc. 
In addition, we can see that the response areas of the five-facial points attributes like nose, 
month and eyes are smaller, while that of the attributes such as smile, mouth opening and hair 
are larger. This phenomenon is also consistent with our common sense. The facial regions involved 
in the five-facial points only concentrate on the five-facial points' locations themselves, but 
there are no fixed areas for the attributes such as hair, smile and so on. Moreover paying 
attention to the attributes of male and bangs, we can see that when the value of the attribute is 0 
(i.e., the label in the figure is black), the attention map does not all turn black, but still 
responds to certain areas. It indicates that whether the value of the attribute is 0 or 1, the 
generated image always has corresponding areas to well reflect its meaning, which demonstrates the effectiveness of the proposed two-path embedding layer.

We can also compare the generated images in SIGM module on condition of with and without SCM 
module in Figure \ref{fig:merge_four}. The second row represents the experiment results without 
the constraints of SCM module in the objective function, and third row represents the results 
of the complete AFGAN model proposed by us. The input 
attribute label is shown in the top color block containing the attribute number and the corresponding content. The ground truth images corresponding to the attribute labels 
are shown in the last row. It can be seen that the quality of the image generated by our proposed 
AFGAN model with SCM module is clearer and more realistic than that of without SCM module.
The second row shows the generated images with two-path embedding layer and self-attention layer, 
and there are some distortion phenomena due to the lack of SCM module. By adding the constraints 
of SCM module, the generated images in third row make progresses in terms of diversity and 
authenticity, and reflect the details of facial texture well.

\subsubsection{ The generated image of three stages in SIGM module}
\begin{figure*}[htbp]
      \centering
      \includegraphics[width=0.8\textwidth]{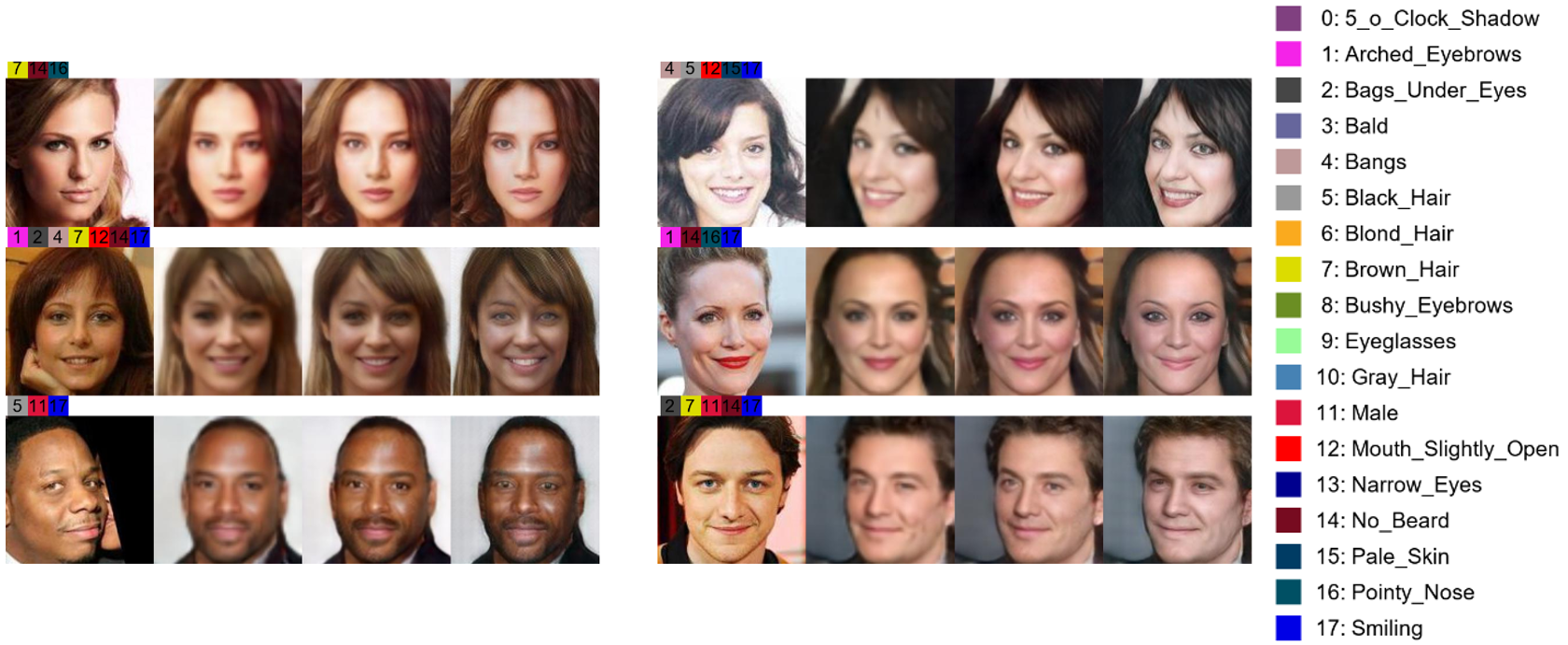}
      \vspace{-2ex}
      \caption{The generated face images of three stages in SIGM module. In each set of four images, the 
      left-most of each group is the ground truth image of the corresponding label, and the three 
      images on the right are $64\times64$, $128\times128$ and $256\times256$ pixels respectively. The images smaller than 
      $256\times256$ pixels are enlarged to $256\times256$ pixels by bilinear interpolation. The input attribute 
      label is shown in the top color block and number, and the corresponding attribute content is 
      listed on the right side.}
      \label{fig:merge_stages}
      \vspace{-3ex}
\end{figure*}
Figure \ref{fig:merge_stages} shows the images generated by three stages of SIGM module separately. 
The images smaller than $256\times256$ pixels are enlarged to $256\times256$ pixels by bilinear interpolation. 
The left-most of each group is the ground truth image with the corresponding label, and the three 
images on the right are $64\times64$, $128\times128$ and $256\times256$ pixels respectively. We can see that the $64\times64$ 
pixel image generated in the first stage mainly reflects the overall attribute of the face, such 
as gender, hairstyle. Then some fine-grained texture information is gradually reflected in the 
following two stages, such as wrinkles of the face, texture of the hair, lightness of the face and 
so on. Moreover, the faces generated in the three stages are consistent. The faces generated in 
latter stage do not change the identity information in the previous stage, but only improve the 
details of the face. In this way, the facial image can be generated step by step in three stages 
according to the input attribute label, which illustrates that SIGM module is effective.
\subsection{ Qualitative evaluations}

We compare the images generated by AttnGAN~\cite{xu2018attngan} and our proposed AFGAN in Figure 
\ref{fig:merge_four}. The images generated by our method are randomly selected. The first row represents the results of AttnGAN model. The second row represents the results of AFGAN model proposed by us. The ground truth images corresponding to the 
attribute labels are shown in the last row. The generated face images by AttnGAN are blurred and have the phenomenon of mode collapse~\cite{goodfellow2014generative}. On the contrary, the 
images generated by AFGAN greatly alleviate the phenomenon of mode collapse and distortion.
\subsection{ Quantitative evaluations}
Moreover, we employ BRISQUE~\cite{mittal2012no}, IS~\cite{salimans2016improved}, FID~\cite{heusel2017gans} and MS-SSIM~\cite{wang2003multiscale} to further evaluate our model quantitatively. 
BRISQUE calculates the no-reference image quality score for images using the Blind/Referenceless Image Spatial Quality Evaluator. A smaller score indicates better perceptual quality. 
IS and FID are typical image evaluation metrics of the GAN model, which respectively focus on the diversity of generated images and the feature distance between the generated images and the real images.
MS-SSIM means multi-scale structural similarity approach for image quality assessment, which provides more flexibility than single-scale approach in incorporating the variations of image resolution and viewing conditions.
For AttnGAN, we use attribute vector instead of text as input. The results are shown in Table~\ref{tab:metric}.
\begin{table}[!hbtp]
\begin{center}
\caption{More quantitative analysis by metric of BRISQUE, IS, FID and MS-SSIM.}
    \label{tab:metric}
    \begin{tabular}{c|c|c|c|c}
        \hline
           & BRISQUE~$\downarrow$ &  IS~$\uparrow$ & FID~$\downarrow$ & MS-SSIM~$\downarrow$ \\ \hline
        AttnGAN~\cite{xu2018attngan} & 62.843 & 5.124 & 40.254 & 0.398 \\
        Wang et al.~\cite{wang2018attribute} & --- & 2.2 & 43.8 & --- \\
        AFGAN(ours) & 35.979 & 5.853 & 36.607 & 0.347 \\
        \hline
    \end{tabular}
    \end{center}
    \vspace{-4ex}
\end{table}

For the metric of BRISQUE, AFGAN achieves 35.979, which is much lower than the result of AttnGAN, i.e., 62.843. It means that the quality of generated images from AFGAN is much better than AttnGAN. 
For the metric of MS-SSIM, AFGAN achieves 0.347 which is much lower than 0.398 obtained by AttnGAN. The lower value of MS-SSIM means the generated images are more diverse. 
For the metric of IS, AFGAN achieves a better result of 5.853 compared to 5.124 achieved by AttnGAN and 2.2 by Wang et al., which demonstrates that the images generated by AFGAN have better image quality again. 
For the metric of FID, AFGAN achieves 36.607 while AttnGAN and Wang et al. achieve 40.254 and 43.8 respectively, which means that the images generated by AFGAN show better authenticity and are closer to the real images. 
All the results show the superiority of our AFGAN over AttnGAN~\cite{xu2018attngan} and Wang et al.~\cite{wang2018attribute} for generating high quality faces.

Besides, we use the CelebA dataset~\cite{liu2015faceattributes} to train an attribute prediction model. Then we 
use the generated images by our model to test the classification accuracy compared to the corresponding input 
facial attributes. The results of AttnGAN, AFGAN without AEM, AFGAN without SCM and AFGAN are shown in Table \ref{tab:class}. As seen, AFGAN achieves the highest classification accuracy, which means that the images generated by AFGAN can 
better reflect the input attributes.
\begin{table}[htbp]
  \vspace{-2ex}
\caption{The classification accuracy of generated facial images by different setting.}
\label{tab:class}
\vspace{-3ex}
    \begin{center}
    \begin{tabular}{|c|c|}
        \hline
        Setting & Classification accuracy \\
        \hline
	     AttnGAN & 0.902   \\   
	     AFGAN w/o AEM  & 0.924   \\
         AFGAN w/o SCM  & 0.940   \\
         AFGAN(ours)  & 0.955   \\
        \hline
    \end{tabular}
  \end{center}
  \vspace{-6ex}
\end{table}

\section{Conclusion}
In this work, we propose a novel attributes aware face image generation method with 
generative adversarial networks called AFGAN to solve the problem of generating 
facial image with specific attributes. AFGAN is composed of three modules, i.e., AEM, SIGM and SCM. AEM uses 
a two-path embedding layer and self-attention layer to convert binary attribute 
vector to rich attribute features. SIGM conducts three stacked generators to generate $64\times64$, 
$128\times128$ and $256\times256$ resolution faces respectively. And SCM propose an 
image-attribute matching loss to enhance the correlation between the generated images 
and input attributes. Both qualitative and quantitative evaluations on CelebA dataset 
show the effectiveness of AFGAN.
\vspace{-1ex}
\section{Acknowledgment}
This work is partially supported by Natural Science Foundation of China (Nos. 61806188, 61976219), and the state key development program in
13th Five-Year under Grant (No. Y808401).






\bibliographystyle{IEEEtran}
\bibliography{ref}
%



\end{document}